\documentclass[11pt]{article}

\usepackage[preprint]{acl}

\usepackage{times}
\usepackage{latexsym}

\usepackage[T1]{fontenc}

\usepackage[utf8]{inputenc}

\usepackage{amsmath}
\usepackage{microtype}
\usepackage{hyperref}
\usepackage{inconsolata}
\usepackage{bbm}
\usepackage{booktabs} 
\usepackage{graphicx}

\title{Cross-Lingual Sentiment Misalignment: Auditing Multilingual Language Models for Inversion Risk, Dialectal Representation, and Affective Stability}

\author{
  Nusrat Jahan Lia \\
  Institute of Information Technology \\
  University of Dhaka, Dhaka, Bangladesh \\
  \texttt{bsse1306@iit.du.ac.bd} \And
  Shubhashis Roy Dipta \\
  University of Maryland, Baltimore County \\
  Baltimore, Maryland, USA \\
  \texttt{sroydip1@umbc.edu}
}
\begin{document}
\maketitle

\begin{abstract}
Recent advances in multilingual representation learning aim to bridge the performance gap between high- and low-resource languages, yet their ability to preserve affective meaning across languages remains underexplored, particularly for underrepresented languages like Bengali. This research addresses cross-lingual sentiment misalignment between Bengali and English by introducing a controlled benchmarking framework evaluating four multilingual transformer models on parallel Bengali-English sentence pairs, stratified by dialect, to assess their representational stability. We demonstrate that a compressed model architecture exhibits a 28.7\% ``Sentiment Inversion Rate,'' fundamentally misinterpreting positive semantics as negative (or vice versa). Consequently, we identify a cross-lingual sentiment skew that we call ``Asymmetric Empathy'', where models systematically dampen or artificially amplify the affective weight of Bengali text relative to its exact English counterpart. Finally, we expose a key vulnerability regarding dialectal representation: a ``Modern Bias'' in the regional model, which exhibits a 57\% increase in alignment error when processing the formal Bengali register compared to modern colloquial text. As foundational encoders continue to serve as safety classifiers and reward models for LLM pipelines, cross-lingual reliability becomes a critical concern. We therefore advocate for the integration of ``Affective Stability'' metrics into future cross-lingual benchmarks to detect and penalize polarity inversions, particularly in low-resource settings.
\end{abstract}

\begin{figure}[t]
    \centering
    \includegraphics[width=1\linewidth]{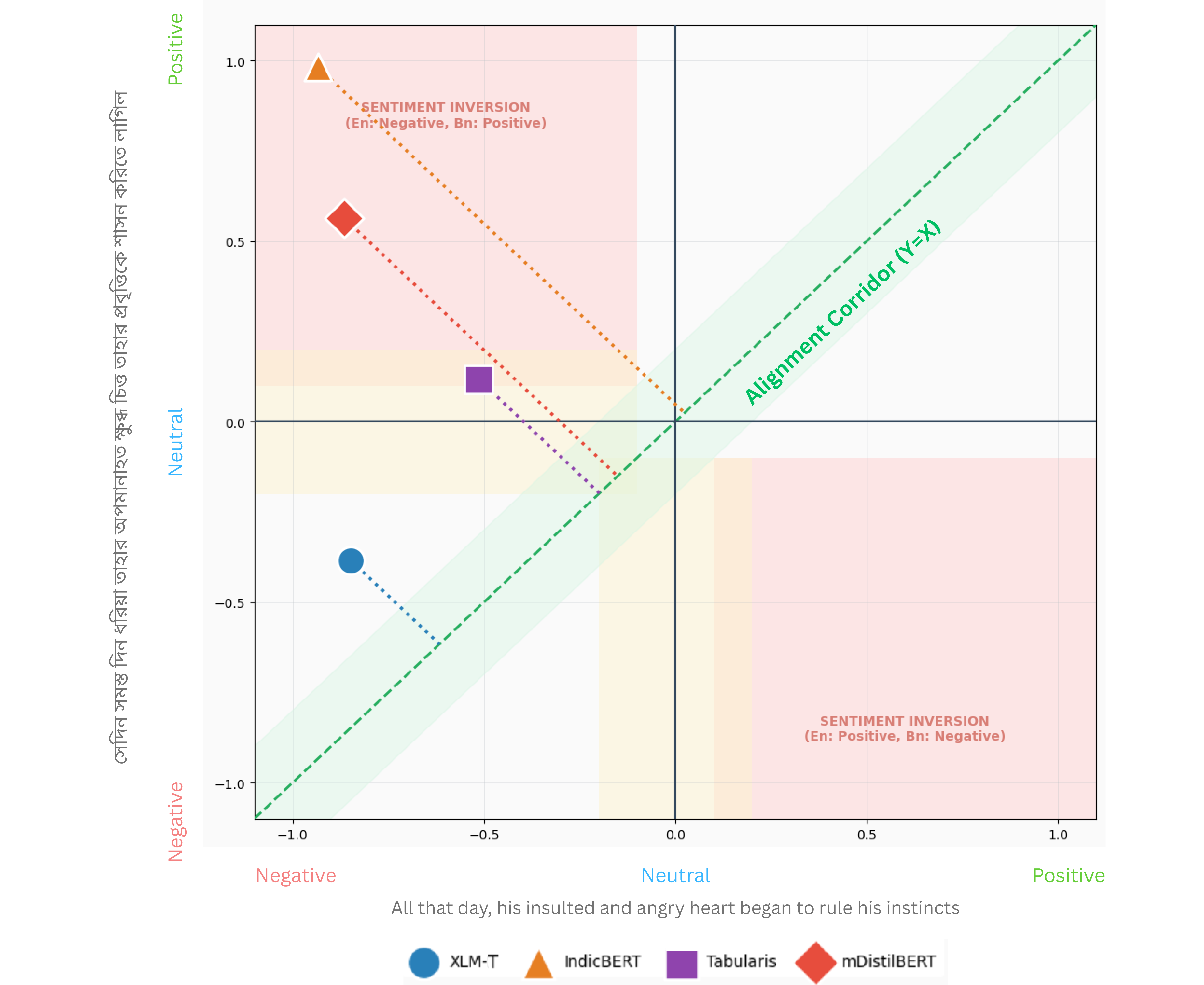}
    \caption{The plot maps the predicted English sentiment vector ($X$-axis) against the Bengali vector ($Y$-axis) for a test sentence. The green corridor represents ideal cross-lingual sentiment alignment ($Y \approx X$). While the large-scale XLM-T model successfully preserves polarity, the compressed (mDistilBERT) and regionally specialized (IndicBERT) architectures are projecting a negative English statement into a positive Bengali latent space (Red Zone).}
    \label{fig:models-on-test-sent}
\end{figure}

\section{Introduction}
Multilingual models have rapidly become the backbone of language-agnostic information access, spanning sentiment analysis, content moderation, retrieval-augmented systems, and downstream knowledge-intensive applications. However, when a multilingual model correctly identifies that an English sentence carries negative sentiment but simultaneously classifies its Bengali semantic equivalent as positive (Figure~\ref{fig:models-on-test-sent}), the representational promise of multilingualism collapses in practice.

Multilingual sentiment encoders, including those for Bengali, are deployed as core components in content moderation systems, where biases can propagate to real-world safety failures; for instance, an audit of Bengali sentiment tools revealed identity-based inconsistencies that undermine moderation accuracy \cite{das2024colonial}. Similarly, \citet{tan2025lionguard} use multilingual classifiers for government-scale moderation in diverse languages with reliance on foundational encoders for LLM safety pipelines. Such instances illustrate the risks of sentiment misalignment in operational safety classifiers.

To mitigate the \textit{curse of multilinguality}, the phenomenon where adding more languages to a fixed-capacity model degrades per-language performance \cite{conneau2020unsupervised}, we investigate the ``Sentiment Inversion'' crisis and its broader representational implications. We demonstrate that the multilingual curse manifests not only as accuracy degradation but also as affective inversion. We employ a standardized cross-lingual sentiment alignment framework to evaluate the semantic consistency of four multilingual sentiment classifier models on parallel Bengali--English text (see Table~\ref{tab:model_map}).

We show that alignment is not uniform across languages or dialects, as summarized in the following key findings:

\textbf{Capacity Constraints \& Alignment Instability:} We demonstrate that the capacity-constrained mDistilBERT architecture exhibits a 28.7\% ``Sentiment Inversion Rate,'' accompanied by a heavy-tailed distribution of errors (Figure~\ref{fig:error_density}), suggesting that model compression may compromise the safety margins required for cross-lingual alignment. We further find that regional specialization alone does not resolve this issue; however, a distilled architecture (Tabularis) that leverages synthetic data can reduce such representational misalignments.

\textbf{Asymmetric Empathy:} We reveal that models exhibit consistent cross-lingual directional bias, systematically dampening or artificially inflating the emotional intensity of non-English text.

\textbf{The Dialectal Sentiment Gap:} We find a ``Modern Bias'' in which models align well with colloquial text but underperform on formal dialects, which are the backbone of Bengali literature.

\section{Related Works}
This section reviews prior work in multilingual representation, cross-lingual sentiment analysis, and dialect-aware NLP, with a focus on gaps in affective alignment and evaluation.
\subsection{Bengali in the Multilingual NLP Landscape}
Bengali remains one of the most underrepresented major languages in NLP despite its global speaker base. \citet{kabir2024benllm} provide a comprehensive audit of LLM performance on Bengali NLP tasks, identifying key failure modes including language generation errors, verbose mismatching with evaluation metrics, and task-specific weaknesses. \citet{bhowmik2025evaluating} document consistent performance gaps for Bengali relative to English across recent LLMs, tracing these to tokenization inefficiency, where models fragment Bengali script into excessive subword units, degrading semantic coherence. The BnMMLU benchmark further demonstrates that even large-scale frontier models show sublinear returns in Bengali reasoning as model size increases, an empirical fingerprint of the multilingual curse at scale \cite{joy2025bnmmlu, conneau2020unsupervised}. 

For sentiment analysis specifically, BanglaBERT \cite{bhattacharjee2022banglabert} established a strong monolingual baseline, and ensemble transformer systems \cite{hoque2024exploring} have achieved high aggregate accuracy. However, in a landmark audit of Bengali sentiment analysis tools, \citet{das2024colonial} reveal that aggregate accuracy conceals systematic identity-based biases, with tools exhibiting differential performance across gender, religious, and national identity signals. This colonial impulse in tool design highlights how reductionist representations reanimate historical hierarchies and motivates the external auditing approach we adopt. Our work extends this critical perspective by focusing specifically on cross-lingual affective misalignment and dialectal representational harm.

\subsection{The Multilingual Curse and Capacity Constraints}
The curse of multilinguality, originally formalized by \citet{conneau2020unsupervised}, describes the empirical observation that, under fixed model capacity, adding more languages to pretraining initially benefits low-resource languages through positive transfer but eventually degrades per-language performance due to inter-language parameter competition. Recent work has substantially refined this understanding. \citet{blevins2024breaking} demonstrate that the curse can be partially lifted through Cross-lingual Expert Language Models (X-ELM), which decouple per-language capacity via modular training and outperform jointly trained multilingual models across 16 languages. \citet{foroutan2025revisiting} further argue that the curse arises not from language count per se but from finite model capacity amplifying the impact of noisy, low-quality data in low-resource languages. This has direct implications for Bengali, where pretraining data is both scarce and noisier than for high-resource languages.

\subsection{Cross-Lingual Sentiment Analysis and Representational Failures}

The adoption of transformer architectures has substantially advanced sentiment analysis in low-resource languages \cite{bhowmick2021sentiment}. Cross-lingual transfer from high-resource to low-resource languages has been enabled both through shared representations in pretrained multilingual models \cite{conneau2020unsupervised} and through machine translation strategies \cite{poncelas2020impact}. \citet{chen2025bridging} document that while GPT-4 achieves approximately 84.4\% F1 in English sentiment, this drops to around 67\% for low-resource languages, and propose adaptive self-alignment strategies with data augmentation to partially close this gap. Recent literature shows that hybrid approaches retaining lexicon features maintain stability advantages over purely neural representations \cite{mahmud2024enhancing}, and ensemble methods can achieve high aggregate accuracy \cite{hoque2024exploring}.

A growing body of literature documents that high aggregate accuracy routinely masks representational failures \cite{das2024colonial}. \citet{wasi2024exploring} show that LLMs acquire social biases through surface linguistic cues, and that Bengali dialectal variation, particularly religious dialect variation, induces systematic performance divergence in large models. \citet{ochieng2025reasoning} extend this critique, demonstrating that reasoning-based LLM sentiment evaluation in low-resource, culturally nuanced contexts reveals failures invisible to label-prediction benchmarks. The CuLEmo benchmark \cite{belay2025culemo} further shows that multilingual LLMs systematically fail to capture culturally grounded variations in emotional expression across languages. These cultural layers of failure are distinct from, but related to, the cross-lingual affective misalignment we document.

\subsection{Bengali Diglossia and Dialectal NLP}

Dialectal variation represents a significant challenge for multilingual NLP, as \citet{wasi2024exploring} demonstrate through empirical evaluation of LLMs on Bengali religious dialects. Bengali exhibits a well-documented diglossic structure comprising Sadhu Bhasha (formal/literary, Sanskrit-derived vocabulary, archaic conjugation) and Cholito Bhasha (colloquial/standard, simplified morphology, contemporary vocabulary), a stylistic split that presents significant challenges for multilingual NLP due to the frequent blending of these forms in everyday communication \cite{ayman2025banglablend}. The critical insight motivating our dialect stratification is that training data for multilingual models is overwhelmingly drawn from contemporary digital sources such as large-scale web crawls, social media, and news corpora \cite{conneau2020unsupervised, kakwani2020indicnlpsuite}, creating a training distribution that is inherently skewed toward the more common, modern Cholito Bengali form \cite{ayman2025banglablend}.

\subsection{Benchmarking Gaps and the Need for Affective Stability Metrics}

Current multilingual benchmarks, including XTREME, XNLI, and their derivatives, evaluate cross-lingual performance on semantic tasks such as natural language inference, question answering (e.g., MLQA, TyDiQA), and named entity recognition (e.g., WikiAnn), with XNLI focusing on entailment classification \cite{hu2020xtreme}. While effective at measuring semantic transfer, these benchmarks largely overlook affective fidelity. Recent efforts such as MMAFFBen \cite{liu2025mmaffben} begin to address this gap by introducing affective evaluation, but comprehensive measurement of sentiment preservation and cross-lingual affective alignment remains limited. \citet{ochieng2025reasoning} explicitly call for benchmarks that measure LLM sentiment in low-resource, culturally nuanced contexts beyond label accuracy. \citet{miah2024multimodal} note that translation-based cross-lingual sentiment approaches can achieve high aggregate accuracy while failing to capture culturally grounded variations in emotional expression, often introducing translation biases in affective intensity. 

We interpret these findings as evidence of directional distortions in how sentiment is mapped across languages, which we formalize as asymmetric empathy. Recent work on multilingual bias evaluation \cite{wasi2024exploring, sadhu2025social} has established that social bias in Bengali LLMs operates across gender and religious lines, but has not examined the cross-lingual affective alignment dimension we investigate. Our proposal for affective stability metrics, which explicitly penalize polarity inversions and dialectal divergence, responds to this benchmarking gap, extending recent work on stability-focused evaluation metrics \cite{atil2024llm} to the multilingual affective domain.

\section{Methodology}
We employ a controlled experimental framework to quantify cross-lingual sentiment alignment in multilingual transformer architectures. We adopt a \textit{within-model} comparative design in which each transformer processes parallel Bengali-English text pairs independently, enabling direct measurement of semantic divergence without inter-model architectural confounds.

\subsection{Dataset Specification}
We utilize a parallel corpus comprising $n = 7{,}350$ Bengali-English sentence pairs, sourced from the publicly available ``BanglaBlend'' dataset \cite{ayman2025banglablend}. Formally, the dataset $\mathcal{D}$ is defined as a set of tuples:

\begin{equation}
\mathcal{D} = \{(B_i, E_i, D_i) \mid i \in [1, n]\}
\end{equation}

\noindent where:
\begin{itemize}
    \item $B_i \in \Sigma_{\text{Bengali}}^*$ represents the $i$-th Bengali sentence (original text),
    \item $E_i \in \Sigma_{\text{English}}^*$ represents the corresponding English translation,
    \item $D_i \in \{\text{Sadhu}, \text{Cholito}\}$ denotes the Bengali dialect classification.
\end{itemize}

Bengali exhibits diglossia with two primary written forms:

\begin{enumerate}
    \item \textbf{Sadhu Bhasha}: Formal/literary register, characterized by Sanskrit-derived vocabulary and archaic verb conjugations.
    
    \item \textbf{Cholito Bhasha}: Colloquial/standard register with relatively simplified morphology and contemporary vocabulary.
\end{enumerate}

\subsection{Models Evaluated}
We benchmark four multilingual transformer architectures representing distinct design paradigms: XLM-T, a large-scale model fine-tuned on high-volume multilingual social media data; IndicBERT, a regionally specialized encoder for Indian languages; Tabularis, a distilled multilingual model enhanced with synthetic data for broad cross-lingual coverage; and mDistilBERT, a compressed multilingual model designed for efficient zero-shot sentiment transfer. Full repository mappings are provided in Table~\ref{tab:model_map}. This selection enables systematic analysis of how scale, regional specialization, synthetic augmentation, and compression interact with cross-lingual affective alignment.

These specific models were selected based on three core inclusion criteria: they are publicly accessible, they are capable of zero-shot Bengali sentiment inference without requiring further task-specific fine-tuning, and together they cover a principled spectrum of architectural capacity (large-scale vs. compressed) and design intent (global vs. regional). Furthermore, because our experimental design isolates cross-lingual representation and affective transfer as the primary variables of interest, we focus exclusively on multilingual architectures, purposefully excluding monolingual models.

\subsection{Experimental Design}

For each model $M$ and sentence pair $(B_i, E_i)$, we perform independent inference on both languages using the same model weights $\theta_M$:

\begin{align}
\hat{y}_{B,i} &= M_\theta(B_i) \quad \text{[Bengali stream]} \\
\hat{y}_{E,i} &= M_\theta(E_i) \quad \text{[English stream]}
\end{align}

Any divergence between $\hat{y}_{B,i}$ and $\hat{y}_{E,i}$ is attributable to cross-lingual representation, calibration, or decision boundary alignment within the same parameter space.

\subsection{Score Normalization and Metric Formulation}

\subsubsection{Universal Score Normalizer}

To enable direct comparison, we define a universal normalization function $\varphi: \text{Predictions} \rightarrow [-1, 1]$:

\begin{equation}
\varphi(\hat{y}) = \begin{cases}
\hat{y}.\text{score} & \text{if } \hat{y}.\text{label} \in \{\text{positive}\} \\
-\hat{y}.\text{score} & \text{if } \hat{y}.\text{label} \in \{\text{negative}\} \\
0 & \text{if } \hat{y}.\text{label} \in \{\text{neutral}\}
\end{cases}
\end{equation}

\noindent \textit{*Note: Standard ``Positive'' and ``Negative'' labels map to $\pm s$ for 2-class and 3-class models, while intermediate positive/negative classes are scaled by 0.5 in the 5-class Tabularis model to accommodate the ``Very Positive'' and ``Very Negative'' extremes. This maintains a uniform linear spacing across sentiment intensity levels, ensuring that the five classes are equidistant on the sentiment continuum.}

After normalization, we obtain continuous sentiment scores:
\begin{align}
S_{B,i} &= \varphi(\hat{y}_{B,i}) \in [-1, 1] \\
S_{E,i} &= \varphi(\hat{y}_{E,i}) \in [-1, 1]
\end{align}

\noindent where $S_{B,i}$ and $S_{E,i}$ denote the Bengali and English sentiment scores, respectively.
\subsubsection{Sentence-Level Alignment Metrics}

For each sentence pair $i$, we compute four alignment metrics:

\paragraph{M1. Alignment Divergence}
\begin{equation}
D_i = |S_{B,i} - S_{E,i}| \in [0, 2]
\end{equation}

\textit{Interpretation:} $D_i = 0.0$ indicates perfect alignment (identical sentiment), $D_i = 0.5$ indicates moderate divergence, $D_i = 2.0$ indicates maximal divergence (opposite extremes). 

\paragraph{M2. Directional Bias}
\begin{equation}
B_i = S_{E,i} - S_{B,i} \in [-2, 2]
\end{equation}

\textit{Interpretation:} $B_i > 0$ indicates that the English text is predicted as more positive (or less negative) than its Bengali counterpart; $B_i < 0$ indicates the reverse; $B_i \approx 0$ indicates minimal cross-lingual divergence for that pair.

\paragraph{M3. Polarity Inversion (Safety Metric)}
\begin{equation}
I_i = \mathbbm{1}\left[
\begin{aligned}
&(S_{B,i} > \tau \land S_{E,i} < -\tau) \\
&\lor (S_{B,i} < -\tau \land S_{E,i} > \tau)
\end{aligned}
\right]
\end{equation}
where $\mathbbm{1}[\cdot]$ is the indicator function and $\tau = 0.1$ is a noise threshold to avoid false positives from near-zero scores.

\textit{Interpretation:} $I_i = 1$ indicates sentiment inversion (e.g., Bengali=Positive, English=Negative); $I_i = 0$ indicates polarity preserved. Inversion is the most severe failure mode, as it indicates misalignment of sentiment direction.

\subsection{Population-Level Aggregation and Statistical Computation}
\label{sec:population_metrics}

To characterize model-level performance and evaluate representational equity across Bengali diglossia, we aggregate the sentence-level metrics into population-level statistics. We compute these across the entire dataset $\mathcal{D}$ as well as its stratified dialectal subsets ($\mathcal{D}_{\text{Sadhu}}$ and $\mathcal{D}_{\text{Cholito}}$).

Table \ref{tab:population_metrics} details the mathematical formulations and interpretations for all population-level evaluation criteria, including overall alignment statistics, safety indicators, and specialized metrics designed to quantify the dialectal gap.

\begin{table}[t]
  \centering
  \small
  \resizebox{\columnwidth}{!}{
  \begin{tabular}{lcccc}
    \toprule
    \textbf{Metric} & \textbf{Tabularis} & \textbf{XLM-T} & \textbf{IndicBERT} & \textbf{mDistilBERT} \\
    \hline
    Mean Div. & 0.200 & 0.276 & 0.375 & 0.417 \\
    Std Dev. & 0.214 & 0.298 & 0.607 & 0.429 \\
    Sadhu Div. & 0.239 & 0.286 & 0.459 & 0.456 \\
    Cholito Div. & 0.161 & 0.266 & 0.292 & 0.379 \\
    Dialect Gap & 0.078 & 0.020 & 0.167 & 0.077 \\
    Sadhu Err. Inc. (\%) & 48.4 & 7.6 & 57.1 & 20.5 \\
    Robustness (\%) & 43.1 & 42.1 & 58.3 & 34.2 \\
    Inversions & 635 & 267 & 1471 & 2107 \\
    Inv. Rate (\%) & 8.6 & 3.6 & 20.0 & 28.7 \\
    Dir. Bias (En-Bn) & 0.002 & 0.057 & 0.106 & -0.066 \\
    \hline
    \end{tabular}}
    \caption{Metric scores per model}
    \label{tab:alignment-gap-full}
\end{table}
\begin{table*}[t]
\centering
\renewcommand{\arraystretch}{1.5}
\small
\resizebox{\textwidth}{!}{%
\begin{tabular}{@{}p{3.1cm} p{6.3cm} p{5.3cm}@{}}
\toprule
\textbf{Metric} & \textbf{Formulation} & \textbf{Interpretation} \\
\midrule
\textbf{Mean Alignment Error} \newline (Divergence) & $\mu_D(M) = \frac{1}{n} \sum_{i=1}^{n} D_i$ & Lower $\mu_D$ indicates better average cross-lingual consistency. \\
\textbf{Std. Deviation} \newline (Divergence) & $\sigma_D(M) = \sqrt{\frac{1}{n-1} \sum_{i=1}^{n} (D_i - \mu_D)^2}$ & Quantifies the variability in alignment quality across the dataset. \\
\textbf{Robustness Index} & $\mathcal{R}(M) = \frac{1}{n} \sum_{i=1}^{n} \mathbbm{1}[D_i < 0.1] \times 100\%$ & \% of pairs with negligible divergence; measures the ``safe operating zone.'' \\
\textbf{Inversion Rate} & $\mathcal{I}_{\text{rate}}(M) = \frac{1}{n} \sum_{i=1}^{n} I_i \times 100\%$ & \% of sentence pairs exhibiting sentiment polarity flips. \\
\textbf{Mean Directional Bias} & $\mu_B(M) = \frac{1}{n} \sum_{i=1}^{n} B_i$ & $>0$: English favored; $<0$: Bengali favored; $\approx0$: No systematic language skew. \\
\midrule
\textbf{Formal Penalty} \newline (Dialect Gap) & $\Delta_{\text{dialect}}(M) = \mu_D(M, \text{Sadhu}) - \mu_D(M, \text{Cholito})$ & $>0$ implies a ``Modern Bias'' where the model struggles with formal registers. \\
\textbf{Relative Dialect Error} & $\Delta_{\text{dialect}}^{\%}(M) = \frac{\Delta_{\text{dialect}}(M)}{\mu_D(M, \text{Cholito})} \times 100\%$ & Normalizes the formal penalty by the baseline colloquial error rate for fair comparison. \\
\bottomrule
\end{tabular}%
}
\caption{Summary of population-level alignment and dialectal metrics.}
\label{tab:population_metrics}
\end{table*}

\section{Results}

Beyond average alignment error, our evaluation identifies three critical failure modes in multilingual model behavior. Table~\ref{tab:alignment-gap-full} presents the corresponding quantitative results.

\subsection{Finding 1: Sentiment Inversion and Alignment Instability}

\begin{figure}[h]
  \centering
  \includegraphics[width=\linewidth]{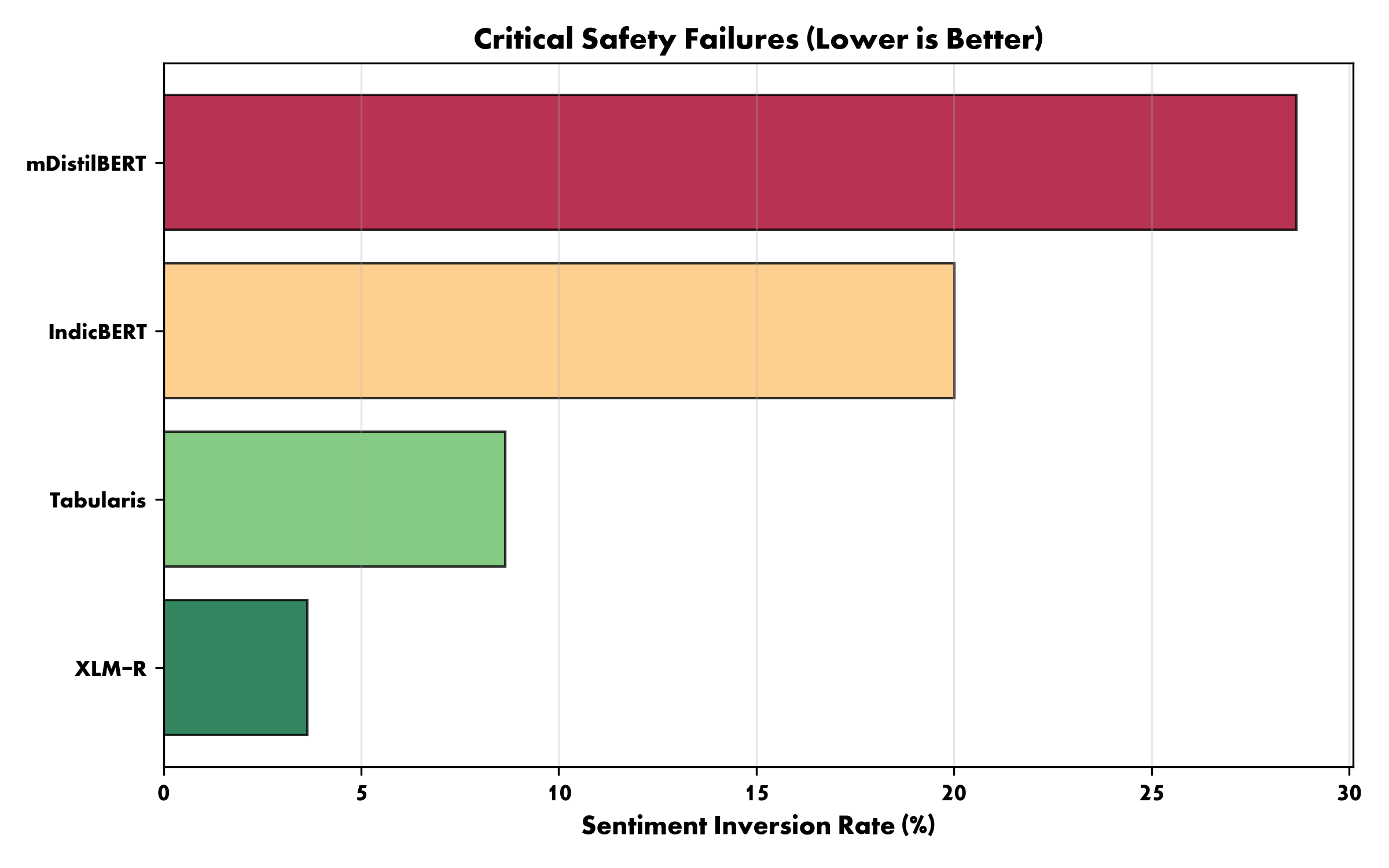}
  \caption{Sentiment Inversion Rate Across Models}
  \label{fig:inversion-pct}
\end{figure}

We define a \textit{sentiment inversion} as a case where a Bengali-English translation pair (similar meaning) receives opposite polarity classifications (positive vs. negative). Such inversions represent major alignment failures, as propositional meaning is preserved but affective interpretation is reversed. Across model architectures, inversion rates vary dramatically, revealing a potential relationship between compression, divergence magnitude, and optimization (see Figure~\ref{fig:inversion-pct}).

\begin{figure}[h]
  \centering
  \includegraphics[width=\linewidth]{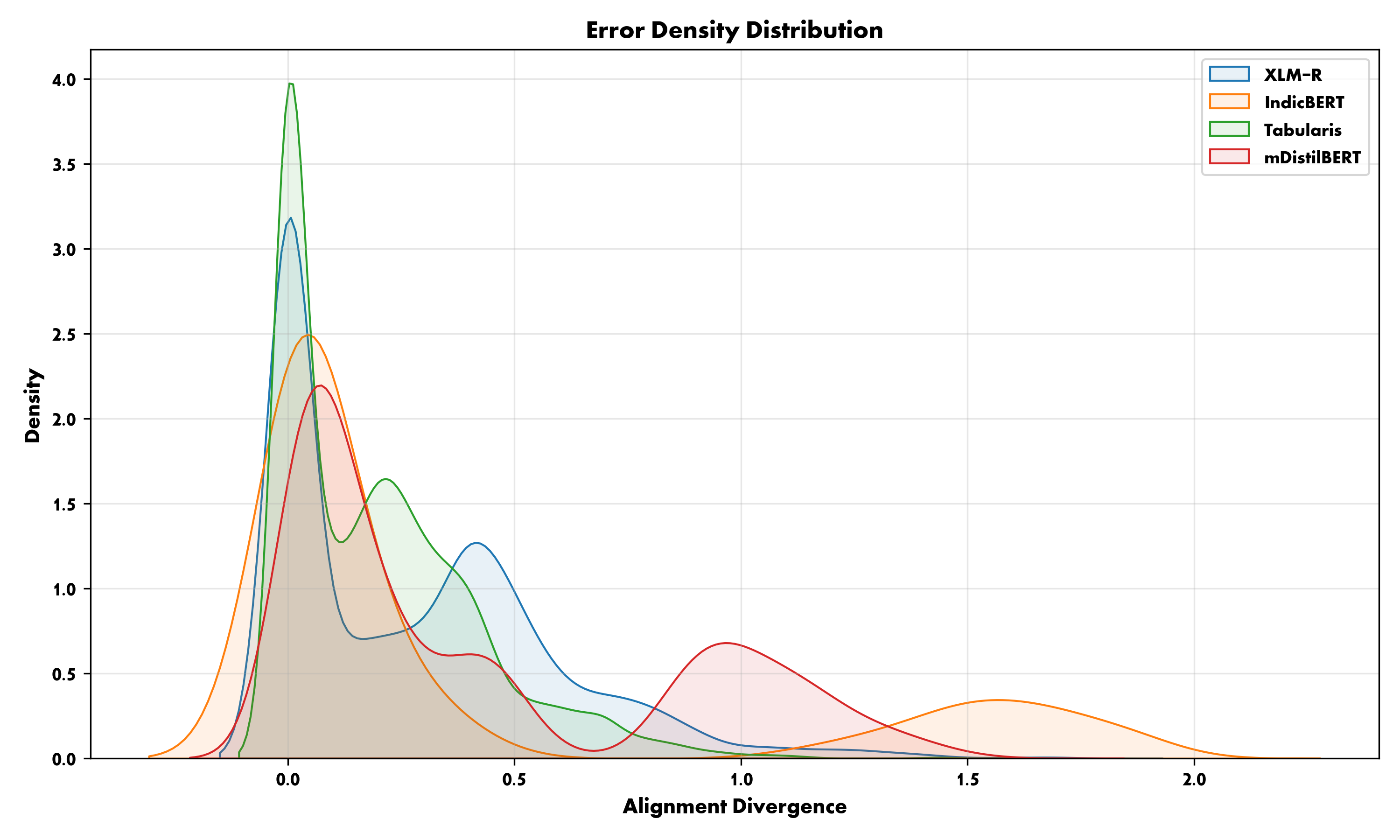}
  \caption{Distribution of Alignment Error Density}
  \label{fig:error_density}
\end{figure}

\begin{itemize}

\item \textbf{Compression and Elevated Inversion Risk.}
The distilled multilingual architecture (mDistilBERT) exhibits the highest mean alignment divergence and lowest robustness (see Table~\ref{tab:alignment-gap-full}). Nearly one in three sentence pairs processed by the compressed model receives directly contradictory affective classifications across Bengali and English. This indicates that while compression improves efficiency, it may disproportionately reduce the representational capacity required for reliable affective calibration. As a result, sentiment polarity reversal emerges as a critical failure mode that distorts core cross-lingual meaning.

\item \textbf{Heavy-Tailed Failure Distribution.}
Alignment error density analysis (Figure~\ref{fig:error_density}) shows that divergence is not normally distributed. Instead, the compressed architecture exhibits a long right tail, corresponding to extreme polarity flips. While IndicBERT achieves the highest robustness metric, it simultaneously exhibits a massive standard deviation in divergence and a high inversion rate with alignment errors that are not normally distributed. This non-normal distribution has practical implications: mean divergence understates actual risk, and models cannot be reliably characterized by their average behavior for deployment in critical downstream applications.

\item \textbf{Scale-Driven Inversion Resilience.}
The large-scale multilingual model (XLM-T) records the lowest polarity inversion rate (Table~\ref{tab:alignment-gap-full}) across all evaluation pairs. This suggests that massive parameter scale and diverse pre-training may preserve coherent affect mappings more effectively than regional specialization, buffering against semantic instability. Crucially, however, the robust distilled Tabularis model (8.6\% inversion rate) implies that compression with data-centric optimization can substantially close the gap between compressed and full-scale architectures. Hence, scale is not the only path to alignment stability. As a DistilBERT-based model fine-tuned with diverse synthetic multilingual data, Tabularis shows that targeted training strategies, rather than scale alone, can drive alignment stability.

\end{itemize}

\subsection{Finding 2: Representational Harm and the Dialectal Gap}

To evaluate robustness under Bengali diglossia, we compare alignment divergence across colloquial (Cholito) and formal (Sadhu) variants. A multilingual system should maintain stable cross-lingual calibration regardless of lexical register. However, we observed a dialectal sensitivity pattern.

\begin{itemize}

\item \textbf{Modern-Register Overfitting.}
Both IndicBERT and Tabularis exhibit sharp increases in divergence when processing formal Sadhu text (see Table~\ref{tab:alignment-gap-full}). This indicates that alignment quality is concentrated in modern and high-frequency lexical distributions, while archaic or formal constructions fall outside the model’s calibrated semantic manifold. We term this phenomenon \textit{Modern Bias}: strong alignment in contemporary usage, but underperformance in formal registers. The \textit{Modern Bias} finding has direct consequences for linguistic equity: users who communicate in formal registers including academic, literary, and administrative Bengali, receive poorer cross-lingual sentiment alignment than users of colloquial varieties. As a result, current multilingual systems risk institutionalizing a structural inequity in which access to reliable inference is contingent on conforming to simplified or non-native linguistic norms.

Conversely, XLM-T demonstrates strong dialectal resilience, likely due to its large-scale multilingual training over diverse text distributions, which provides broader lexical and syntactic coverage and supports stable affective mappings across regional variation.

\end{itemize}

\subsection{Finding 3: Asymmetric Empathy and Directional Bias}

In a multilingual architecture, the system must preserve not just the polarity but the intensity of user intent. We evaluate this using \textit{Directional Bias} (English score $-$ Bengali score). An ideally aligned system should yield a distribution centered at zero with low variance. Instead, we observed two distinct alignment regimes (as seen in Table~\ref{tab:alignment-gap-full}, Figure~\ref{fig:directional_bias} and Figure~\ref{fig:comparative-analysis})

\begin{itemize}
\item \textbf{Compression-Induced Bengali Positivity Skew.}
The distilled architecture (mDistilBERT) exhibits a negative directional bias, indicating that it systematically scores Bengali text as more positive (or less negative) than its exact English translation. For a human user in a safety-critical context: the model may artificially dampen the severity of a negative Bengali sentiment, underweighting the severity of negative Bengali content relative to equivalent English content. As demonstrated in the case study (Figure~\ref{fig:comparative-analysis}), mDistilBERT exhibits a safety failure by correctly scoring an English statement as deeply negative (-0.981) while assigning a positive score (+0.533) for its exact Bengali equivalent.

\item \textbf{Regional English Optimism Bias.}
Conversely, IndicBERT demonstrates a positive mean directional bias, assigning higher positivity (or lower negativity) to English inputs relative to their Bengali counterparts. This imposes an equity penalty on Bengali users, as their neutral or moderately negative statements are penalized with harsher negative classifications compared to English speakers.

\end{itemize}

These findings indicate that cross-lingual affective misalignment is consistently and directionally structured rather than stochastic, which can lead to downstream distortions in applications relying on Bengali sentiment signals.

\subsection{Proposed Metric: Affective Stability Index}
\label{sec:affective_stability}

Our empirical evaluation reveals that aggregate alignment error ($\mu_{D}$) alone is insufficient to capture the critical safety failures inherent in cross-lingual representation. Specifically, models with relatively moderate average divergence may still exhibit high rates of sentiment inversion ($\mathcal{I}_{rate}$) (Table~\ref{tab:alignment-gap-full}). To address this benchmarking gap and formally quantify cross-lingual reliability, we introduce the Affective Stability Index ($\mathcal{AS}$). We define Affective Stability as a composite metric that rewards tight semantic alignment while strictly penalizing polarity inversions. Utilizing the population-level metrics defined in Section~\ref{sec:population_metrics}, it is computed as:

\begin{equation}
    \mathcal{AS}(M) = \left( 1 - \frac{\mu_{D}(M)}{2} \right) \times (1 - \mathcal{I}_{rate}(M))
\end{equation}

The first term normalizes the Mean Alignment Divergence ($\mu_{D}$) into a similarity score bounded by $[0, 1]$, as the maximum theoretical divergence in our normalized space is 2.0. The second term acts as a strict penalty mask based on the Inversion Rate ($\mathcal{I}_{rate}$), expressed as a probability. An $\mathcal{AS}$ score of 1.0 indicates perfect cross-lingual affective fidelity, whereas lower scores reflect compounding representational misalignments.
\clearpage
Table~\ref{tab:affective_stability} presents the Affective Stability scores for all evaluated architectures. The results validate our observations regarding scale and compression. While the large-scale XLM-T maintains high affective stability, mDistilBERT suffers a degradation in overall reliability. Notably, the distilled Tabularis model achieves an $\mathcal{AS}$ score highly competitive with XLM-T. This empirically demonstrates that targeted training strategies and synthetic data utilization can preserve affective calibration even under capacity constraints.

\begin{table}[h]
\centering
\resizebox{\columnwidth}{!}{
\begin{tabular}{lccc}
\toprule
\textbf{Model} & \textbf{Mean Div. ($\mu_{D}$)} & \textbf{Inv. Rate ($\mathcal{I}_{rate}$)} & \textbf{Affective Stability ($\mathcal{AS}$)} \\
\midrule
XLM-T & 0.276 & 0.036 & \textbf{0.831} \\
Tabularis & 0.200 & 0.086 & 0.823 \\
IndicBERT & 0.375 & 0.200 & 0.650 \\
mDistilBERT & 0.417 & 0.287 & 0.564 \\
\bottomrule
\end{tabular}}
\caption{Affective Stability ($\mathcal{AS}$) of evaluated models.}
\label{tab:affective_stability}
\end{table}

\section{Discussion: Alignment Under Linguistic Pluralism}

Our findings instantiate the multilingual curse at an affective, rather than purely accuracy-based, level. The inversion gap we observe is consistent with \citet{blevins2024breaking}'s theoretical framing: fixed model capacity induces inter-language parameter competition that degrades per-language representation. We hypothesize that compression amplifies this competition in ways that may disrupt the fine-grained representational signals required for stable sentiment polarity mapping. 

Our results suggest two distinct intervention points for multilingual model development. First, at the pretraining data level: the \textit{Modern Bias} finding points to a gap in training corpus composition for Bengali, even when utilizing synthetic data. Formal Sadhu text is underrepresented in web-crawled multilingual corpora (dominated by social media and news in contemporary registers), and correcting this imbalance would directly address dialectal representational harm. Second, at the post-distillation fine-tuning level: our finding that Tabularis substantially outperforms mDistilBERT despite both being distilled architectures demonstrates that targeted fine-tuning with synthetic multilingual corpora can preserve affective calibration through compression. Addressing both stages is particularly vital in low-resource settings like Bengali, where computationally efficient, compressed multilingual models must still deliver equitable and reliable affective understanding across all linguistic communities.

The multilingual NLP evaluation landscape currently lacks standardized metrics for cross-lingual affective consistency. Existing benchmarks \cite{hu2020xtreme, ruder2021xtreme, han2025mubench, goldman2025eclektic} primarily measure semantic and syntactic transfer but do not penalize polarity inversions or account for affective fidelity. We propose that multilingual benchmarks incorporate the metric suite introduced in this work (Inversion Rate, Affective Stability) as standard evaluation dimensions. This is especially critical for downstream applications that depend on affective signals: content moderation, mental health monitoring, customer feedback analysis, and social listening systems operating across languages all face systematic failure risks due to sentiment inversion.

We further argue that the choice is not binary between scale and alignment; targeted distillation strategies, data augmentation, and curated metrics that explicitly prioritize affective calibration can simultaneously address efficiency constraints and ensure representational equity, particularly in low-resource deployment settings.

\section{Conclusion}
We present a cross-lingual sentiment alignment audit comparing four transformer architectures on Bengali-English parallel data stratified by dialect. Our findings reveal that current multilingual models exhibit structured affective representational failures: sentiment inversion under compression, dialectal bias against formal registers, and directional asymmetry in emotional intensity calibration. These failures are distinct from accuracy degradation measured by standard benchmarks and constitute specific threats to equitable language technology access for Bengali users.
We argue that building better multilingual representations requires evaluative interventions that make affective alignment failures visible. Future multilingual benchmarks should incorporate Inversion Rate and Affective Stability as standard dimensions. Future training and alignment research should address the formal register gap in Bengali pretraining data and explore dialect-stratified post-training alignment as a path to equitable compression. We believe these directions are generalizable beyond Bengali to the broader landscape of low-resource and dialectally diverse languages underrepresented in current multilingual NLP infrastructure.

\bibliography{custom}

@article{ayman2025banglablend,
  title={BanglaBlend: A large-scale nobel dataset of bangla sentences categorized by saint and common form of bangla language},
  author={Ayman, Umme and Saha, Chayti and Rahat, Azmain Mahtab and Khushbu, Sharun Akter},
  journal={Data in Brief},
  volume={58},
  pages={111240},
  year={2025},
  publisher={Elsevier}
}

@inproceedings{das2024colonial,
  title={The``Colonial Impulse" of Natural Language Processing: An Audit of Bengali Sentiment Analysis Tools and Their Identity-based Biases},
  author={Das, Dipto and Guha, Shion and Brubaker, Jed R and Semaan, Bryan},
  booktitle={Proceedings of the 2024 CHI Conference on Human Factors in Computing Systems},
  pages={1--18},
  year={2024}
}

@article{chen2025bridging,
  title={Bridging resource gaps in cross-lingual sentiment analysis: adaptive self-alignment with data augmentation and transfer learning},
  author={Chen, Li and Shang, Shifeng and Wang, Yawen},
  journal={PeerJ Computer Science},
  volume={11},
  pages={e2851},
  year={2025},
  publisher={PeerJ Inc.}
}

@article{hoque2024exploring,
  title={Exploring transformer models in the sentiment analysis task for the under-resource Bengali language},
  author={Hoque, Md Nesarul and Salma, Umme and Uddin, Md Jamal and Ahamad, Md Martuza and Aktar, Sakifa},
  journal={Natural Language Processing Journal},
  volume={8},
  pages={100091},
  year={2024},
  publisher={Elsevier}
}

@article{wasi2024exploring,
  title={Exploring bengali religious dialect biases in large language models with evaluation perspectives},
  author={Wasi, Azmine Toushik and Islam, Raima and Islam, Mst Rafia and Rafi, Taki Hasan and Chae, Dong-Kyu},
  journal={arXiv preprint arXiv:2407.18376},
  year={2024}
}

@article{ochieng2025reasoning,
  title={Reasoning Beyond Labels: Measuring LLM Sentiment in Low-Resource, Culturally Nuanced Contexts},
  author={Ochieng, Millicent and Thieme, Anja and Ezeani, Ignatius and Ueno, Risa and Maina, Samuel and Ronen, Keshet and Gonzalez, Javier and O'Neill, Jacki},
  journal={arXiv preprint arXiv:2508.04199},
  year={2025}
}

@inproceedings{poncelas2020impact,
  title={The impact of indirect machine translation on sentiment classification},
  author={Poncelas, Alberto and Lohar, Pintu and Hadley, James and Way, Andy},
  booktitle={Proceedings of the 14th Conference of the Association for Machine Translation in the Americas (Volume 1: Research Track)},
  pages={78--88},
  year={2020}
}

@inproceedings{bhowmick2021sentiment,
  title={Sentiment analysis for Bengali using transformer based models},
  author={Bhowmick, Anirban and Jana, Abhik},
  booktitle={Proceedings of the 18th International Conference on Natural Language Processing (ICON)},
  pages={481--486},
  year={2021}
}

@article{mahmud2024enhancing,
  title={Enhancing sentiment analysis in bengali texts: A hybrid approach using lexicon-based algorithm and pretrained language model bangla-bert},
  author={Mahmud, Hemal and Mahmud, Hasan and Rashid, Mohammad Rifat Ahmmad},
  journal={arXiv preprint arXiv:2411.19584},
  year={2024}
}

@inproceedings{conneau2020unsupervised,
  title={Unsupervised cross-lingual representation learning at scale},
  author={Conneau, Alexis and Khandelwal, Kartikay and Goyal, Naman and Chaudhary, Vishrav and Wenzek, Guillaume and Guzm{\'a}n, Francisco and Grave, Edouard and Ott, Myle and Zettlemoyer, Luke and Stoyanov, Veselin},
  booktitle={Proceedings of the 58th annual meeting of the association for computational linguistics},
  pages={8440--8451},
  year={2020}
}

@inproceedings{bhattacharjee2022banglabert,
  title={BanglaBERT: Language model pretraining and benchmarks for low-resource language understanding evaluation in Bangla},
  author={Bhattacharjee, Abhik and Hasan, Tahmid and Ahmad, Wasi and Mubasshir, Kazi Samin and Islam, Md Saiful and Iqbal, Anindya and Rahman, M Sohel and Shahriyar, Rifat},
  booktitle={Findings of the Association for Computational Linguistics: NAACL 2022},
  pages={1318--1327},
  year={2022}
}

@inproceedings{kakwani2020indicnlpsuite,
  title={IndicNLPSuite: Monolingual corpora, evaluation benchmarks and pre-trained multilingual language models for Indian languages},
  author={Kakwani, Divyanshu and Kunchukuttan, Anoop and Golla, Satish and NC, Gokul and Bhattacharyya, Avik and Khapra, Mitesh M and Kumar, Pratyush},
  booktitle={Findings of the association for computational linguistics: EMNLP 2020},
  pages={4948--4961},
  year={2020}
}

@article{bhowmik2025evaluating,
  title={Evaluating LLMs' Multilingual Capabilities for Bengali: Benchmark Creation and Performance Analysis},
  author={Bhowmik, Shimanto and Dipto, Tawsif Tashwar and Islam, Md Sazzad and Hsu, Sheryl and Reasat, Tahsin},
  journal={arXiv preprint arXiv:2507.23248},
  year={2025}
}

@inproceedings{kabir2024benllm,
  title={BenLLM-eval: A comprehensive evaluation into the potentials and pitfalls of large language models on Bengali NLP},
  author={Kabir, Mohsinul and Islam, Mohammed Saidul and Laskar, Md Tahmid Rahman and Nayeem, Mir Tafseer and Bari, M Saiful and Hoque, Enamul},
  booktitle={Proceedings of the 2024 Joint International Conference on Computational Linguistics, Language Resources and Evaluation (LREC-COLING 2024)},
  pages={2238--2252},
  year={2024}
}

@article{joy2025bnmmlu,
  title={BnMMLU: Measuring Massive Multitask Language Understanding in Bengali},
  author={Joy, Saman Sarker and Shatabda, Swakkhar},
  journal={arXiv preprint arXiv:2505.18951},
  year={2025}
}

@inproceedings{blevins2024breaking,
  title={Breaking the curse of multilinguality with cross-lingual expert language models},
  author={Blevins, Terra and Limisiewicz, Tomasz and Gururangan, Suchin and Li, Margaret and Gonen, Hila and Smith, Noah A and Zettlemoyer, Luke},
  booktitle={Proceedings of the 2024 conference on empirical methods in natural language processing},
  pages={10822--10837},
  year={2024}
}

@article{foroutan2025revisiting,
  title={Revisiting multilingual data mixtures in language model pretraining},
  author={Foroutan, Negar and Teiletche, Paul and Tarun, Ayush Kumar and Bosselut, Antoine},
  journal={arXiv preprint arXiv:2510.25947},
  year={2025}
}

@inproceedings{belay2025culemo,
  title={CULEMO: Cultural lenses on emotion-benchmarking LLMs for cross-cultural emotion understanding},
  author={Belay, Tadesse Destaw and Ahmed, Ahmed Haj and Ii, Alvin C Grissom and Ameer, Iqra and Sidorov, Grigori and Kolesnikova, Olga and Yimam, Seid Muhie},
  booktitle={Proceedings of the 63rd Annual Meeting of the Association for Computational Linguistics (Volume 1: Long Papers)},
  pages={18894--18909},
  year={2025}
}

@article{miah2024multimodal,
  title={A multimodal approach to cross-lingual sentiment analysis with ensemble of transformer and LLM},
  author={Miah, Md Saef Ullah and Kabir, Md Mohsin and Sarwar, Talha Bin and Safran, Mejdl and Alfarhood, Sultan and Mridha, Md F},
  journal={Scientific Reports},
  volume={14},
  number={1},
  pages={9603},
  year={2024},
  publisher={Nature Publishing Group UK London}
}

@article{atil2024llm,
  title={LLM Stability: A detailed analysis with some surprises},
  author={Atil, Berk and Chittams, Alexa and Fu, Liseng and Ture, Ferhan and Xu, Lixinyu and Baldwin, Breck},
  journal={arXiv preprint arXiv:2408.04667},
  volume={1},
  year={2024}
}

@article{liu2025mmaffben,
  title={Mmaffben: a multilingual and multimodal affective analysis benchmark for evaluating LLMs and VLMS},
  author={Liu, Zhiwei and Qian, Lingfei and Xie, Qianqian and Huang, Jimin and Yang, Kailai and Ananiadou, Sophia},
  journal={arXiv preprint arXiv:2505.24423},
  year={2025}
}

@inproceedings{hu2020xtreme,
  title={Xtreme: A massively multilingual multi-task benchmark for evaluating cross-lingual generalisation},
  author={Hu, Junjie and Ruder, Sebastian and Siddhant, Aditya and Neubig, Graham and Firat, Orhan and Johnson, Melvin},
  booktitle={International conference on machine learning},
  pages={4411--4421},
  year={2020},
  organization={PMLR}
}

@inproceedings{sadhu2025social,
  title={Social bias in large language models for bangla: An empirical study on gender and religious bias},
  author={Sadhu, Jayanta and Saha, Maneesha Rani and Shahriyar, Rifat},
  booktitle={Proceedings of the First Workshop on Language Models for Low-Resource Languages},
  pages={204--218},
  year={2025}
}

@misc{tabularisai2025multilingualsentiment,
  author    = {Vadim Borisov and Samuel Gyamfi and Richard H. Schreiber},
  title     = {Multilingual Sentiment Analysis},
  year      = {2025},
  doi       = {10.57967/hf/5968},
  url       = {https://huggingface.co/tabularisai/multilingual-sentiment-analysis},
  publisher = {Hugging Face},
  note      = {Revision 69afb83}
}

@article{goldman2025eclektic,
  title={Eclektic: a novel challenge set for evaluation of cross-lingual knowledge transfer},
  author={Goldman, Omer and Shaham, Uri and Malkin, Dan and Eiger, Sivan and Hassidim, Avinatan and Matias, Yossi and Maynez, Joshua and Gilady, Adi Mayrav and Riesa, Jason and Rijhwani, Shruti and others},
  journal={arXiv preprint arXiv:2502.21228},
  year={2025}
}

@article{han2025mubench,
  title={MuBench: Assessment of Multilingual Capabilities of Large Language Models Across 61 Languages},
  author={Han, Wenhan and Zhang, Yifan and Chen, Zhixun and Liu, Binbin and Lin, Haobin and Zhang, Bingni and Wang, Taifeng and Pechenizkiy, Mykola and Fang, Meng and Zheng, Yin},
  journal={arXiv preprint arXiv:2506.19468},
  year={2025}
}

@inproceedings{ruder2021xtreme,
  title={XTREME-R: Towards more challenging and nuanced multilingual evaluation},
  author={Ruder, Sebastian and Constant, Noah and Botha, Jan and Siddhant, Aditya and Firat, Orhan and Fu, Jinlan and Liu, Pengfei and Hu, Junjie and Garrette, Dan and Neubig, Graham and others},
  booktitle={Proceedings of the 2021 Conference on Empirical Methods in Natural Language Processing},
  pages={10215--10245},
  year={2021}
}

@inproceedings{barbieri2022xlm,
  title={XLM-T: Multilingual language models in Twitter for sentiment analysis and beyond},
  author={Barbieri, Francesco and Anke, Luis Espinosa and Camacho-Collados, Jose},
  booktitle={Proceedings of the thirteenth language resources and evaluation conference},
  pages={258--266},
  year={2022}
}

@inproceedings{tan2025lionguard,
  title={LIONGUARD 2: Building Lightweight, Data-Efficient \& Localised Multilingual Content Moderators},
  author={Tan, Leanne and Chua, Gabriel and Ge, Ziyu and Lee, Roy Ka-Wei},
  booktitle={Proceedings of the 2025 Conference on Empirical Methods in Natural Language Processing: System Demonstrations},
  pages={264--285},
  year={2025}
}

\appendix

\section{Supplementary Figures and Tables}
\label{app:supplementary}
This appendix contains the visualizations and table referenced in the main findings of the paper.

\begin{table}[!h]
\centering
\small
\begin{tabular}{@{}p{1.7cm} p{5.3cm}}
\toprule
\textbf{Model Name} & \textbf{Repository (HuggingFace)} \\
\midrule
XLM-T &
\href{https://huggingface.co/cardiffnlp/twitter-XLM-Toberta-base-sentiment}{\texttt{cardiffnlp/XLM-Toberta-sentiment}} \cite{barbieri2022xlm} \\
\midrule

IndicBERT &
\href{https://huggingface.co/ai4bharat/IndicBERTv2-alpha-SentimentClassification}{\texttt{ai4bharat/IndicBERTv2-sentiment}} \\
\midrule

Tabularis &
\href{https://huggingface.co/Tabularisai/multilingual-sentiment-analysis}{\texttt{tabularisai/multilingual-sentiment}} \\
\midrule

mDistilBERT &
\href{https://huggingface.co/lxyuan/distilbert-base-multilingual-cased-sentiments-student}{\texttt{lxyuan/distilbert-multilingual}} \\
\bottomrule
\end{tabular}
\caption{Model Repository Mapping}
\label{tab:model_map}
\end{table}
\vspace{1ex}
\small \textbf{Note on Tabularis:} This model is a fine-tuned version of the model \texttt{distilbert/distilbert-base-multilingual-cased} for multilingual sentiment analysis. It utilizes synthetic data from multiple sources to achieve robust performance across different languages and cultural contexts \cite{tabularisai2025multilingualsentiment}.

\begin{figure}[!h]
  \centering
  \includegraphics[width=\linewidth]{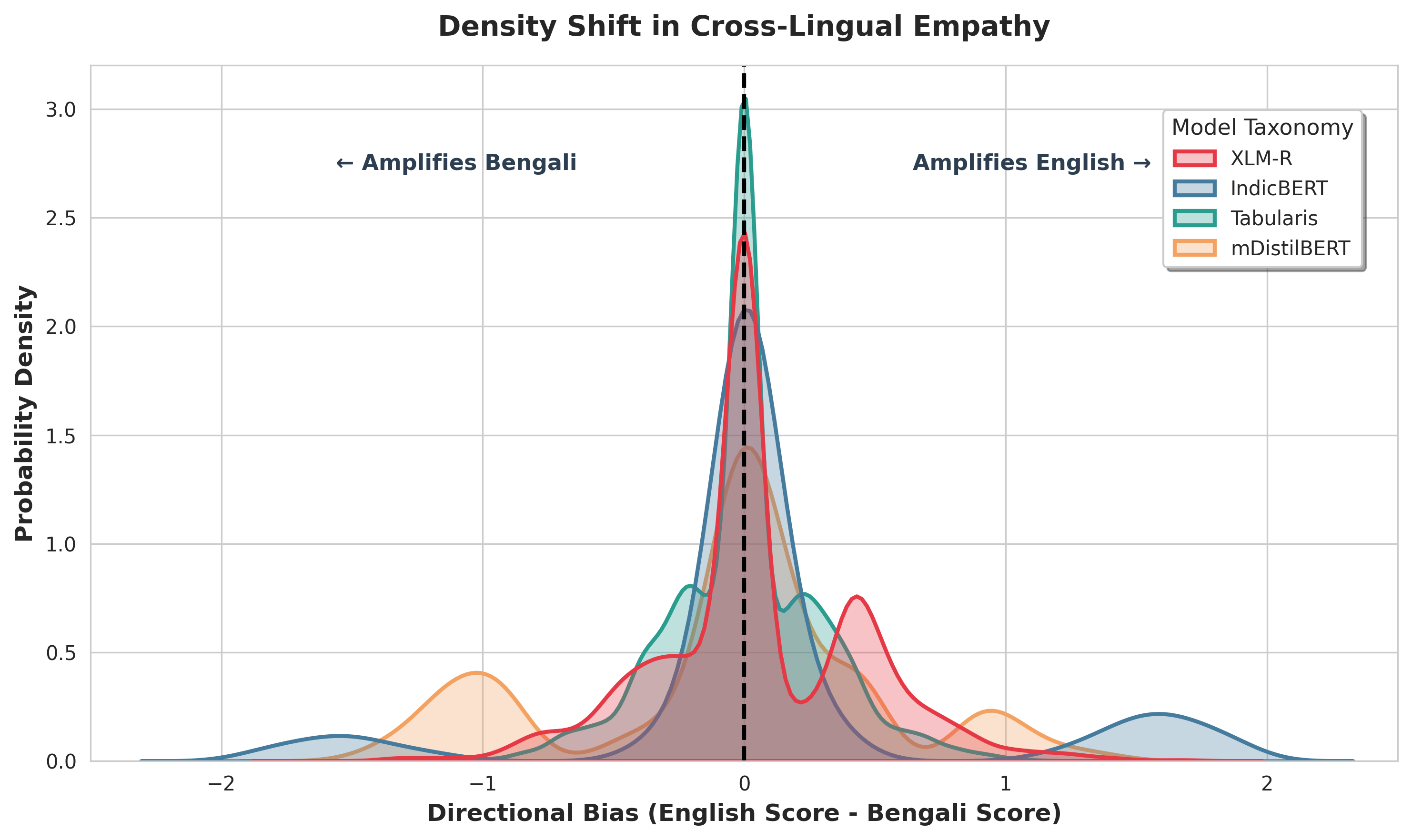}
  \caption{Directional Bias in Sentiment Scores (English $-$ Bengali)}
  \label{fig:directional_bias}
\end{figure}

\begin{figure}[!h]
  \centering
  \includegraphics[width=\linewidth]{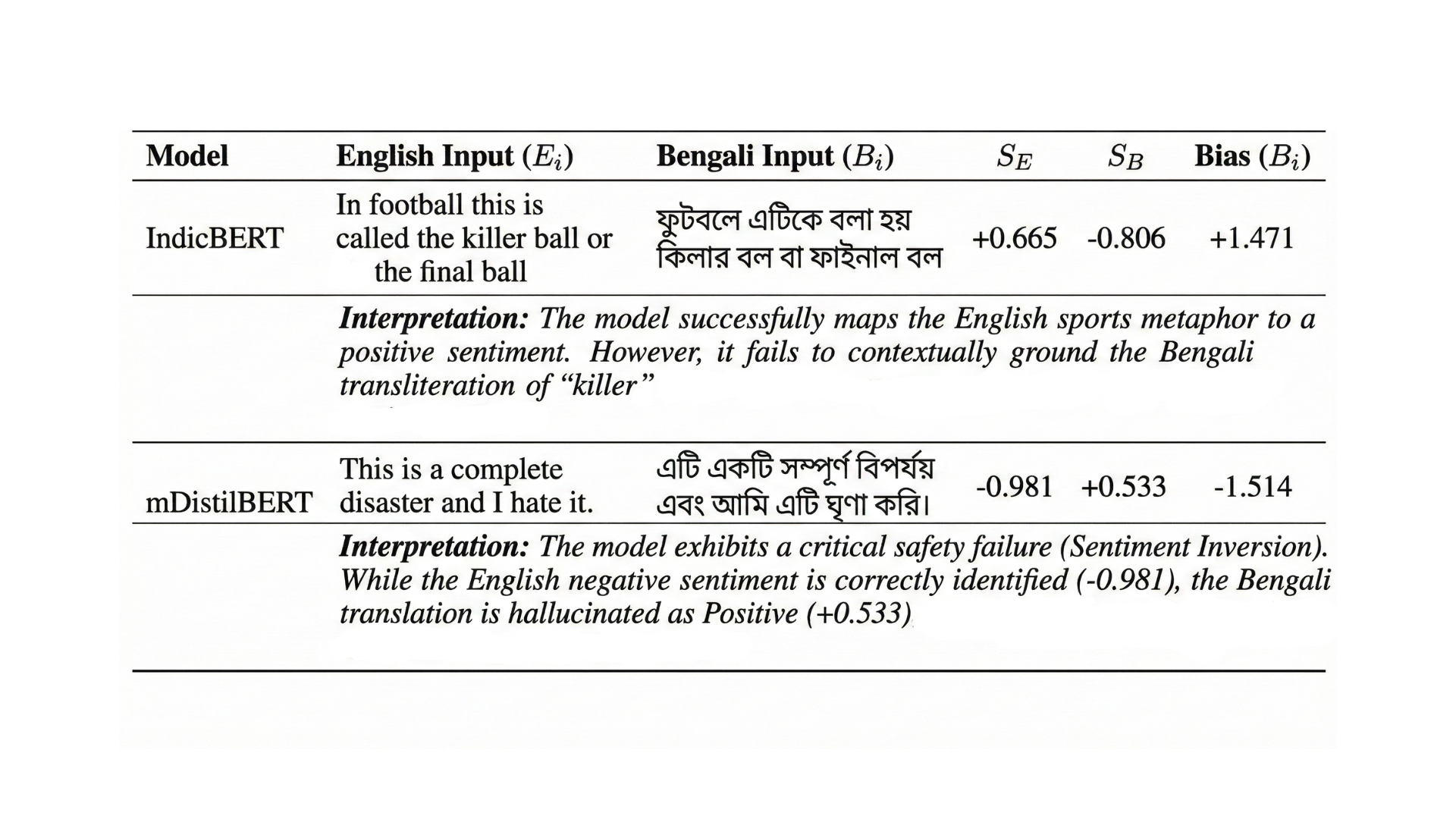}
  \caption{Illustrative case study validating Asymmetric Empathy in cross-lingual sentiment alignment, demonstrating severe instance-level directional bias.}
  \label{fig:comparative-analysis}
\end{figure}

\end{document}